\documentclass{article}

\usepackage[nonatbib, preprint]{neurips_2020}

\usepackage[utf8]{inputenc} %
\usepackage[T1]{fontenc}    %
\usepackage{hyperref}       %
\usepackage{url}            %
\usepackage{booktabs}       %
\usepackage{amsfonts}       %
\usepackage{nicefrac}       %
\usepackage{microtype}      %
\usepackage{algorithm}
\usepackage[noend]{algorithmic}
\usepackage{graphicx}
\usepackage{comment}
\usepackage{wrapfig}
\usepackage{amsmath}
\usepackage{marginnote}
\usepackage{xcolor}

\hypersetup{
 colorlinks=True,
 linkcolor=blue,
 citecolor=blue,
 urlcolor=blue}

\title{Behavioral Priors and Dynamics Models: Improving Performance and Domain Transfer in Offline RL}

\author{
Catherine Cang$^1$ \hspace*{5pt} 
Aravind Rajeswaran$^{2, 3}$ \hspace*{5pt}
Pieter Abbeel$^1$ \hspace*{5pt}
Michael Laskin$^1$ \\[5pt]
\texttt{\{catherinecang, pabbeel, mlaskin\}@berkeley.edu, aravraj@fb.com}\\[5pt]
$^1$ UC Berkeley \hspace*{5pt} $^2$ Facebook AI Research \hspace*{5pt} $^3$ University of Washington
}

\begin{document}

\maketitle

\vspace*{-10pt}

\begin{abstract}
Offline Reinforcement Learning (RL) aims to extract near-optimal policies from imperfect offline data without additional environment interactions. Extracting policies from diverse offline datasets has the potential to expand the range of applicability of RL by making the training process safer, faster, and more streamlined. We investigate how to improve the performance of offline RL algorithms, its robustness to the quality of offline data, as well as its generalization capabilities. To this end, we introduce Offline Model-based RL with Adaptive Behavioral Priors (MABE). Our algorithm is based on the finding that dynamics models, which support within-domain generalization, and behavioral priors, which support cross-domain generalization, are complementary. When combined together, they substantially improve the performance and generalization of offline RL policies. In the widely studied D4RL offline RL benchmark, we find that MABE achieves higher average performance compared to prior model-free and model-based algorithms. In experiments that require cross-domain generalization, we find that MABE outperforms prior methods. Additional details can be found on our \href{https://sites.google.com/berkeley.edu/mabe}{website.}
\end{abstract}

\section{Introduction}

 \begin{wrapfigure}{r}{0.5\textwidth}
     \centering
     \vspace*{-15pt}
       \includegraphics[width=.45\textwidth]{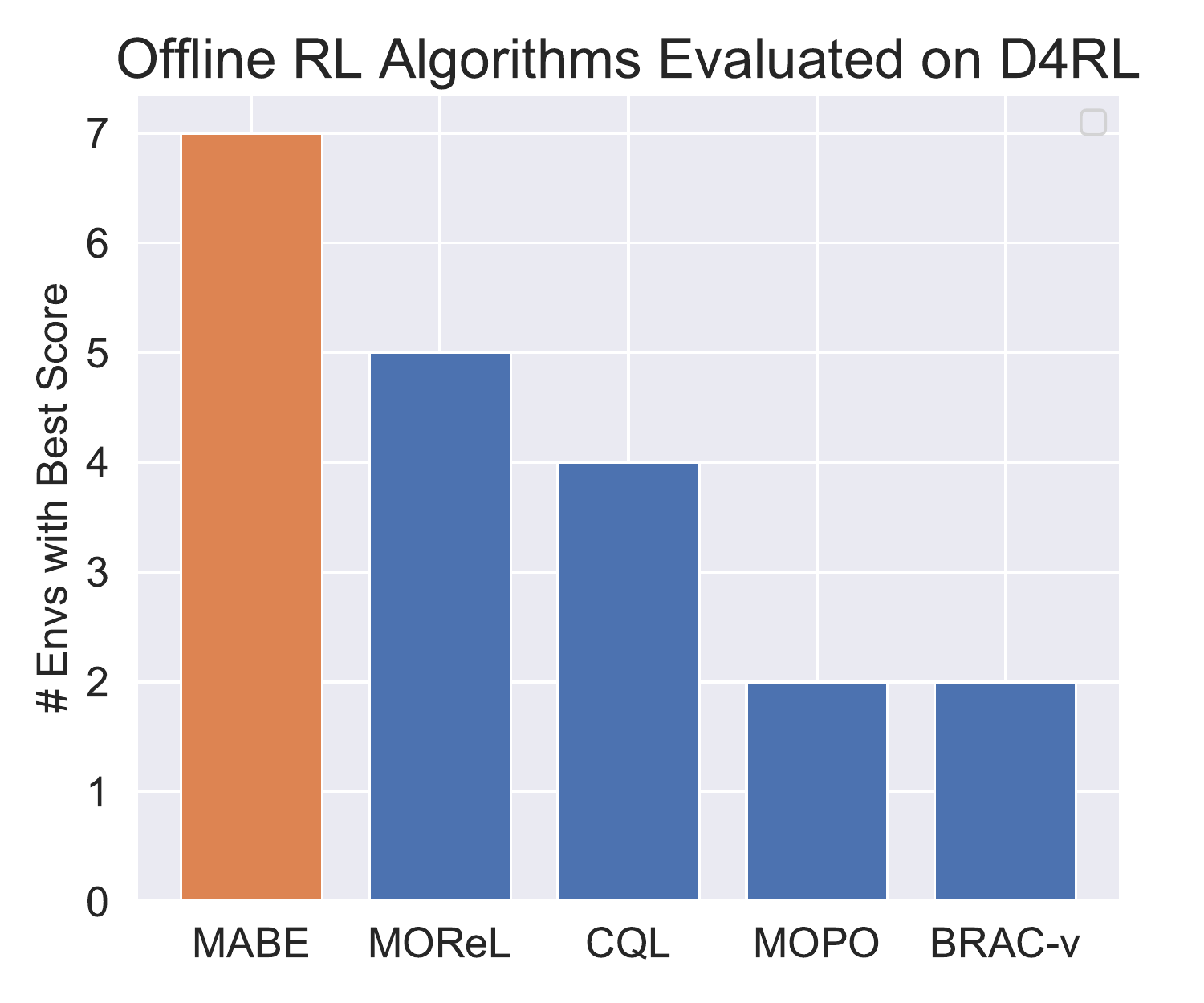}
     \caption{Our proposed algorithm, MABE, when compared to prior work, achieves the top score in {\bf 7 out of 9} D4RL datasets~\cite{fu2020d4rl} we study. We consider multiple algorithms to achieve the top score if they are within 2\% points of each other.}
    \vspace{-5pt}
     \label{fig:flare_intro}
 \end{wrapfigure}

Over the last five years advances in Deep Reinforcement Learning (RL) have been at the source of a number of impressive results in autonomous control, including the ability to solve video games from pixels~\cite{mnih2015human}, master the game of Go~\cite{silver2017mastering}, play multi-agent large scale video games~\cite{vinyals2019alphastar}, and control robots~\cite{OpenAI2019SolvingRC}. Most advances in RL were achieved in simulated environments where data was cheap to collect and mistakes during policy training were harmless. However, two substantial problems stand in the way from utilizing the above approaches to deploy RL algorithms in real-world settings. First, since RL algorithms require millions and sometimes billions of environment interactions, learning policies with RL in the real world is costly in terms of time and resources. Second, since RL algorithms stochastically explore their environment, the resulting agents are not safe and can harm the environment, themselves, or other agents if trained in the real world. How can we overcome the challenges of data efficiency and safety to enable RL algorithms that can be deployed in real world settings?

Offline or Batch RL~\cite{LangeGR12, levine2020offlinerlsurvey} has recently been proposed as a promising paradigm to tackle these challenges. Offline RL agents use logged or previously collected data by humans or other agents for learning. Importantly, the offline data does not have to consist of expert demonstrations like in the case of imitation learning~\cite{pomerleau1988alvinn, Abbeel2004ApprenticeshipLV, Ziebart2008MaximumEI}, but can be collected with policies that are sub-optimal or noisy. Such policies may already be in deployment for a variety of applications like autonomous driving, warehouse automation, dialogue systems~\cite{jaques2019way, ZhouSRE17} and recommendation systems~\cite{CovingtonAS16, SwaminathanJ15}. By learning policies only using offline datasets and perhaps fine-tuning the policy using a small dataset of subsequent interactions, offline RL has the potential to be highly sample efficient and safe.
The primary challenge with extracting policies from offline data comes from the distribution mismatch between transitions seen during training and those encountered during evaluation. Conservatism or pessimism has emerged as a core principle in offline RL to deal with distribution mismatch. Conservatism encourages the offline RL agent to improve the policy while also staying close to the dataset distribution, thereby minimizing distribution shift between training and deployment. A number of algorithms, both model-free and model-based, have been proposed that incorporate conservatism in various forms like importance weights~\cite{LiuSAB19}, value functions~\cite{kumar2019bear, kumar20cql, fujimoto2018addressing, Agarwal2020AnOP}, and dynamics models~\cite{Kidambi-MOReL-20, yu20mopo, ArgensonMBOP, MatsushimaBREMEN}. 

Recently, model-based offline RL algorithms like MOReL~\cite{Kidambi-MOReL-20} and MOPO~\cite{yu20mopo} have demonstrated impressive results in benchmark tasks and also the ability to re-purpose the learned dynamics model to solve downstream tasks that are different from those encountered in the offline dataset. They incorporate conservatism in the learning process by learning pessimistic dynamics models using uncertainty quantification. However, uncertainty quantification with deep neural networks can pose challenges in many domains, such as those with high dimensional input-output spaces or multiple confounding factors~\cite{Ovadia2019CanYT, Begoli2019TheNF, Jiang2018ToTO, Abdar2020ARO, Ribeiro2016LIME}. Since offline RL views uncertainty quantification as a means to the end of incorporating conservatism, and since uncertainty quantification by itself can be a difficult exercise, we are motivated to develop offline RL algorithms that do not require uncertainty quantification. In this work, we develop an algorithm that achieves this goal. Our algorithm outperforms prior approaches in the widely studied D4RL benchmark~\cite{fu2020d4rl} as well as in tasks that require domain adaptation and generalization. Thus, our algorithm has potentially wider applicability, especially in settings where uncertainty estimation can be difficult.

\begin{figure}[t!]
  \centering
       \includegraphics[width=\textwidth]{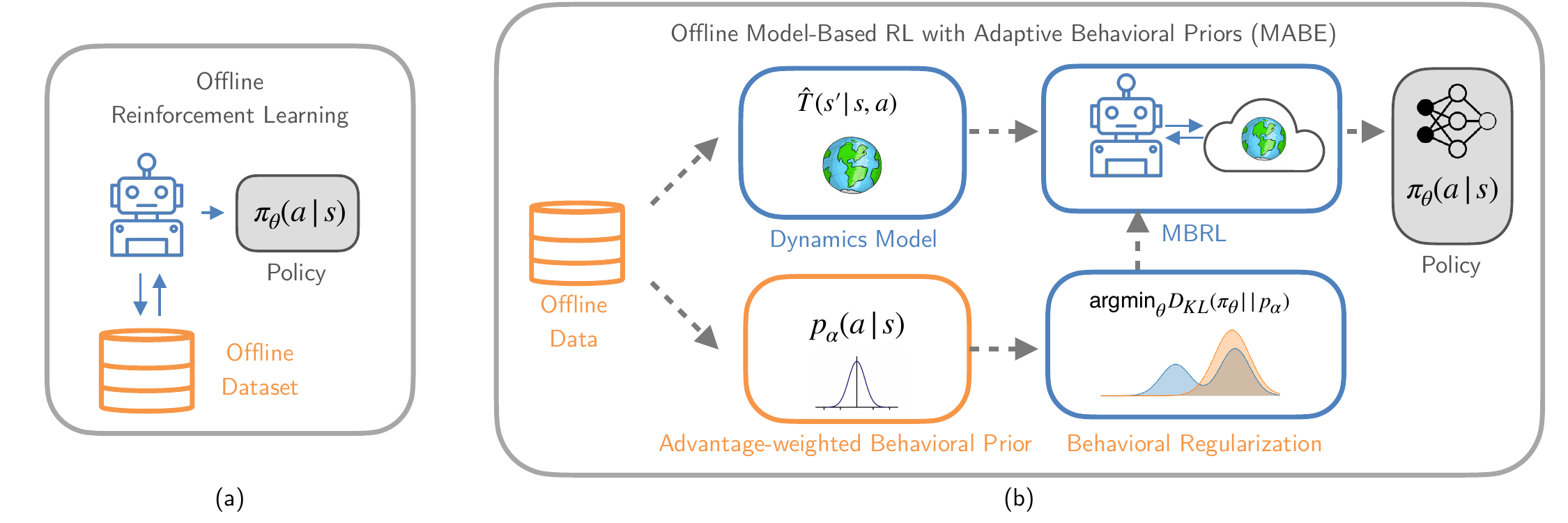}
  \vspace*{-10pt}
  \caption{(a) The offline RL paradigm. Rather than interacting with the environment directly, an agent extracts a policy from an offline dataset. (b) A schematic architecture of our proposed algorithm (MABE). First, by interaction with the the offline dataset, the agent learns a dynamics model and an advantage weighted behavioral prior. Then, the dynamics model generates a synthetic dataset which is used alongside the original offline data to train the policy $\pi_\theta$. Finally, the behavioral prior regularizes the learned policy to keep the agent within the support of the original dataset. }
  \label{fig:mujoco}
  \vspace*{-10pt}
\end{figure}

\paragraph{Our Contribution} Our principal contribution in this work is the development of a new algorithm -- offline model-based RL with Adaptive Behavioral Regularization (MABE). Using the offline dataset, MABE learns an approximate dynamics model, reward function, as well as an adaptive behavioral prior. By adaptive behavioral prior, we mean a policy that approximates the behavior in the offline dataset while giving more importance to trajectories with high rewards. Using the learned dynamics model and reward function, MABE performs model-based RL with an objective to maximize the rewards along with a KL-divergence penalty that encourages the agent to stay close to the adaptive behavioral prior. This divergence penalty provides the necessary conservatism needed to succeed in offline RL. Our major findings in this work are listed below.

\begin{enumerate}
    \itemsep0em
    \item Our algorithm, MABE, achieves the highest scores in 7 out of 9 D4RL~\cite{fu2020d4rl} benchmark tasks we study, as well as the highest average normalized score.
    \item MABE is flexible and can benefit from uncertainty estimation if available or forgo it altogether. Our empirical ablations suggest that uncertainty estimation contributes only minor improvements compared to the other components of dynamics models and behavioral priors. Thus, MABE can be used in a wider set of application domains, especially those where uncertainty estimation is difficult.
    \item We demonstrate that MABE has favorable generalization capabilities to  new tasks by leveraging the learned dynamics model and transferring of behavioral priors across datasets, a capability that is only possible when both model-based and behavioral priors are combined. 
\end{enumerate}

\section{Preliminaries}

We operate in the standard RL setting of infinite horizon discounted Markov Decision Process (MDPs), defined as the tuple $\mathcal{M} = (\mathcal{S},\mathcal{A}, R, T, \rho_0, \gamma)$. The MDP tuple has states $s \in \mathcal{S}$, actions $a\in \mathcal{A}$, rewards $r = R(s,a)$, transition dynamics $s' \sim T(\cdot | s,a)$, an initial state distribution $s_0 \sim \rho_0(\cdot)$, and a discount factor $\gamma \in [0, 1)$. A policy defines a mapping from states to actions, typically in the form of a probability distribution: $a \sim \pi (\cdot | s)$. The value $V^\pi(s)$ and action-value function $Q^\pi(s,a)$ describe the long term reward behavior of policy $\pi$.
\begin{align*}
    V^\pi(s) := \mathbb{E}_{\mathcal{M}, \pi} \left[ \sum_{t=0}^\infty \gamma^t R(s_t,a_t) \mid s_0=s \right], \hspace*{5pt}
    Q^\pi(s, a) := R(s,a) + \gamma \mathbb{E}_{s' \sim T(\cdot|s,a)} \left[ V^\pi(s') \right]
\end{align*}
where the first expectation $\mathbb{E}_{\mathcal{M}, \pi}$ denotes actions are sampled according to $\pi$ and future states are sampled according to the MDP dynamics $T(\cdot|s,a)$. The goal in RL is the learn the optimal policy:
\begin{equation}
\pi^* \in \arg \max_\pi \ J(\pi, \mathcal{M}) := \mathbb{E}_{s \sim \rho_0} \left[ V^\pi(s) \right].
\end{equation}
When the MDP (especially $T$) is unknown, exploration is important to learn the optimal policy.

\paragraph{Model-Based RL (MBRL)} is an approach to learning in MDPs that involves learning an approximate MDP $\widehat{\mathcal{M}} = (\mathcal{S}, \mathcal{A}, \widehat{R}, \widehat{T}, \widehat{\rho_0}, \gamma)$. The learned MDP has the same state and action spaces, but uses the learned approximate dynamics and reward models. Generating samples from $\widehat{\mathcal{M}}$ is cheap and does not require environment interaction. As a result, various algorithms based on policy gradient and dynamic programming~\cite{sutton1998introduction} can be used to efficiently improve the policy, with intermittent data collection to improve model approximation quality. Recently, MBRL algorithms have demonstrated strong results in a variety of RL tasks~\cite{RajeswaranGameMBRL, janner2019mbpo, hafner2019dream, schrittwieser2019mastering}, including offline RL~\cite{Kidambi-MOReL-20, yu20mopo, ArgensonMBOP}.

\paragraph{Offline RL} is a setting in RL where we must learn a policy using a fixed dataset of environment interactions. Specifically, we are given a dataset of interactions $\mathcal{D} = \{ s_i, a_i, s'_i, r_i \}_{i=1}^N$ of $N$ environment interactions collected using one or more behavioral policies. If the behavioral policies do not induce sufficient exploration, it is not possible to learn an optimal policy for the underlying MDP even as $N \rightarrow \infty$~\cite{Chen2019InformationTheoreticCI, Kidambi-MOReL-20}. Thus, the goal in offline RL is typically to learn the best possible policy using the provided dataset.

\paragraph{Model-Based Offline RL} algorithms like MOPO~\cite{yu20mopo} and MOReL~\cite{Kidambi-MOReL-20} leverage MBRL to learn in the offline RL setting. They learn an approximate MDP using the offline dataset. Simulation with the learned MDP allows the offline RL agent to ask counterfactual questions about actions that are unseen in the dataset by leveraging the generalization capabilities of the learned dynamics model. However, since the model cannot be iteratively refined or improved like in the case of online RL, the learned MDP is likely erroneous on out-of-distribution states. As a result, policy learning in the learned MDP may exploit the errors in the model to optimize rewards, leading to poor performance in the true MDP. To guard against this exploitation, MOPO and MOReL penalize the agent for visiting out-of-distribution states in the learned MDP, with uncertainty in the dynamics model being used to detect out-of-distribution states.

\section{Model-Based Offline RL with Adaptive Behavioral Regularization}
\label{sec:method}

Given an offline dataset $\mathcal{D}$, our goal is to learn a parameterized policy $\pi_{\theta}$ that achieves high rewards, without any additional interaction with the environment. We assume $\mathcal{D}$ consists of $\{s, a, s', r\}$ tuples which we use to learn $\hat{T}$ along with a behavioral prior $p_{\alpha}(a_t | s_t)$. This dataset can be collected using one or more structured behavioral policies interacting with test environment. We now present our algorithm MABE (Model-Based Offline RL with Adaptive Behavioral Regularization), which consists of three components described below.

\paragraph{ Dynamics Model Learning} MABE is a model-based RL algorithm, and thus we use the offline dataset to learn a neural network dynamics model. This can be accomplished using maximum likelihood estimation or other generative modeling techniques such as variational models~\cite{hafner2019dream}. Let $\hat{T}_\psi (\cdot | s,a)$ represent the generative model for the conditional next state distribution. Similar to prior offline MBRL works~\cite{yu20mopo, Kidambi-MOReL-20, ArgensonMBOP, MatsushimaBREMEN}, we learn the generative dynamics model with maximum-likelihood learning as:
\begin{equation}
    \max_{\hat{T}_\psi (\cdot | s,a)} \ \ \mathbb{E}_{(s, a, s') \sim \mathcal{D}} \left[ \log \left( \hat{T}_\psi(s' | s,a) \right) \right].
\end{equation}

\paragraph{ Learning Behavioral Priors} Our main insight is the use of adaptive behavioral priors as a form of regularization in offline MBRL. Building on prior work~\cite{brac2019wu,awr2019peng}, we utilize behavioral regularization within the MBRL framework. Our experimental results suggest that combining MBRL with behavioral regularization can incorporate sufficient conservatism to succeed in offline RL. This is in contrast to prior offline MBRL works that rely crucially on uncertainty estimation which may prove difficult in various applications.

We consider a parameterized generative model $p_\alpha (a | s)$ that represents our behavioral prior. A straightforward option is to learn a behavior model that replicates the statistics in the dataset.
\begin{equation}
    \label{eq:unweighted_prior}
    p^{\text{eq}}_{\alpha} \  \in \ \arg \max_{p_\alpha} \ \mathbb{E}_{\tau \sim \mathcal{D}} \left[ \sum_{t=0}^{|\tau|} \log \Big( p_\alpha (a_t | s_t) \Big) \right]
\end{equation}

Alternatively, we can consider an adaptive behavioral prior that is biased towards trajectories that achieve higher rewards. This can be particularly useful in diverse datasets collected with multiple policies -- some of which perform better at the task while other policies may exhibit behaviors that may hinder the task we want the offline RL agent to learn. Similar to Siegel et al.~\cite{Siegel2020KeepDW}, we seek a behavioral prior that is biased towards the high reward trajectories in the dataset while also staying close to the average statistics in the dataset. We formulate this as:
\begin{equation}
\begin{split}
    p^{\text{adapt}}_{\alpha} \  \in \ & \arg \max_{p_\alpha} \ \mathbb{E}_{\tau \sim \mathcal{D}} \left[ \sum_{t=0}^{|\tau|} \omega(s_t, a_t) \cdot p_\alpha(a_t | s_t) \right] \\
    & \text{subject to } \ \mathbb{E}_{s \sim \mathcal{D}} \left[ D_{KL} \left( p_\alpha \| \bar{p} \right) \right] \leq \delta,
\end{split}
\end{equation}
where $\bar{p}$ denotes the empirical behavioral policy and $\omega(s_t, a_t)$ is the weighting function. The non-parametric solution to the above optimization is given by:
\[
p^{\text{adapt}}_{\alpha} (a_t | s_t) \propto \bar{p}(a_t | s_t) \cdot \exp \left( \omega(t, \tau) / \eta \right),
\]
where we have used $\propto$ to avoid specification of the normalization factor, and $\eta$ represents a temperature parameter that is related to the constraint level $\delta$. The above non-parametric policy can be projected into the space of parametric neural network policies as~\cite{awr2019peng, Siegel2020KeepDW}:
\begin{equation}
    \label{eq:adaptive_prior}
    p^{\text{adapt}}_{\alpha} \in \ \arg \max_{p_\alpha} \ \mathbb{E}_{\tau \sim \mathcal{D}} \left[ \sum_{t=0}^{|\tau|} \exp \left( \omega(s_t, a_t) / \eta \right) \cdot \log \Big( p_\alpha (a_t | s_t) \Big) \right].
\end{equation}
For the choice of the weighting function, we use
\[
\omega(s_t, a_t) := \hat{Q}(s_t, a_t) \cdot (1-\gamma) / r_{\max},
\]
where $\hat{Q}$ is learned using TD-error minimization and $r_{\max}$ is the maximum reward observed in the dataset.
In this process, we treat the temperature $\eta$ as the hyper-parameter choice. This implicitly defines the constraint threshold $\delta$, and makes the problem specification and optimization more straightforward.

\paragraph{ Behavior Regularized Model-Based RL} Equipped with a dynamics model and adaptive behavioral prior, our algorithm MABE, performs model-based RL with a regularized objective given by:
\begin{equation}
    \label{eq:mabe_objective}
    \max_{\pi_\theta} \ \underset{(s,a) \sim \rho^{\pi_\theta}_{\hat{\mathcal{M}}}}{\mathbb{E}} \left[ \tilde{r}(s,a) \right], \ \text{ with } \ \tilde{r}(s,a) := \hat{r}_\psi(s,a) - \beta \left( \log \pi_\theta(a | s) + \log p_\alpha(a|s) \right).
\end{equation}
We use $\rho^{\pi_\theta}_{\hat{\mathcal{M}}}$ to denote the discounted state visitation distribution induced by executing $\pi_\theta$ in the learned MDP model. This objective encourages the agent to increase the rewards along with entropy and behavioral regularization. We learn a policy to solve this optimization using SAC~\cite{haarnoja2018soft}, resulting in an algorithm that is similar to a behavior regularized version of Dyna~\cite{sutton1991dyna} and MBPO~\cite{janner2019mbpo}. Algorithm~\ref{alg:mabe} presents the full details of our learning approach.

\begin{algorithm}[h!]
\label{alg:mabe}
\caption{MABE: Model-Based Offline RL with Adaptive Behavioral Regularization}
\begin{algorithmic}[1]
  \STATE \textbf{Inputs:} Offline dataset $\mathcal{D}$, learned dynamics and reward models $\hat{T}_\psi$ and $\hat{r}_\psi$, adaptive behavioral prior $p_{\alpha}(a_t | s_t)$, target divergence $\delta$, learning rates $\lambda_{\pi}$, $\lambda_{Q}$, $\lambda_{\beta}$, $\lambda_\phi$, rollout length $h$
  \STATE Initialize policy $\pi_{\theta}$, critic $Q_{\phi}$, target network $Q_{\bar{\phi}}$, and $\mathcal{D}_\text{aug} = \mathcal{D}$
  \FOR{$N$ epochs}{
    \FOR{$K$ trajectories}{
        \STATE Collect rollouts $(s_t, a_t, r_t, s_{t+1})$ of length $h$ using $\hat{T}_\psi$ and $\hat{r}_\psi$ starting from a randomly chosen state from the offline dataset.
        \STATE $\mathcal{D}_\text{aug} \leftarrow \mathcal{D}_\text{aug} \cup (s_t, a_t, r_t, s_{t+1})$
    }
    \ENDFOR
    \FOR{each gradient step}{
        \STATE Sample a batch of $(s_t, a_t, r_t, s'_t)$ tuples from $\mathcal{D}_\text{aug}$
        \STATE $\bar{Q}=r(s_{t}, a_{t})+\gamma[Q_{\bar{\phi}}(s'_{t}, \pi_{\theta}(\cdot \mid s'_{t}))-\beta D_{\mathrm{KL}}(\pi_{\theta}(\cdot \mid s'_{t}), p_{\alpha}(\cdot \mid s'_{t}))]$
        \STATE $\theta \leftarrow \theta-\lambda_{\pi} \nabla_{\theta}\left[Q_{\phi}(s_{t}, \pi_{\theta}(\cdot \mid s_{t}))-\beta D_{\mathrm{KL}}(\pi_{\theta}(\cdot \mid s_{t}), p_{\alpha}(\cdot \mid s_{t}))\right]$
        \STATE $\phi \leftarrow \phi-\lambda_{Q} \nabla_{\phi}\left[\frac{1}{2}(Q_{\phi}(s_{t}, a_{t})-\bar{Q})^{2}\right]$
        \STATE $\beta \leftarrow \beta - \lambda_{\beta} \cdot \left( D_{\mathrm{KL}}(\pi_{\theta}(\cdot \mid s_{t}), p_{\alpha}(\cdot \mid s_{t}))-\delta \right)$
        \STATE $\bar{\phi} \leftarrow \lambda_\phi \phi+ (1-\lambda_\phi) \bar{\phi}$
    }
    \ENDFOR
  }
  \ENDFOR
\end{algorithmic}
\end{algorithm}

\paragraph{Optional use of uncertainty quantification} MABE is a flexible framework that can additionally incorporate uncertainty quantification if available, in addition to the behavioral prior regularization. Let $u(s,a) \geq D_{TV}(\hat{T}(\cdot|s,a), T(\cdot|s,a))$ be an estimate of the dynamics model uncertainty in state $(s,a)$. Analogous to prior work like MOPO and MOReL, we can additionally incorporate uncertainty into the MABE objective given by Eq.~\ref{eq:mabe_objective} as:
\[
\tilde{r}(s,a) = \hat{r}_\psi(s,a) - \beta \left( \log \pi_\theta(a | s) + \log p_\alpha(a|s) \right) - \xi u(s,a).
\]
We emphasize again that additional reward penalty based on uncertainty is optional, and our experiment results suggest that it only offers marginal benefits compared to our other components.

\section{Results}

\paragraph{MABE design choices} We first outline the main decision choices and implementation details used for our experiments. Our implementation of MABE is built on MOPO. We parameterize the policy, behavioral prior, and dynamics model as a Gaussian distributions, with the mean being parameterized by an MLP network, and the covariance is also learned. For example, the dynamics is represented as
\[
\hat{T}_\psi(s' | s,a) = \mathcal{N} \left( \mathrm{MLP}_\psi(s, a), \ \Sigma_\psi \right).
\]
The reward and Q-function are modeled using deterministic MLP networks. We learn the policy and Q-function using MBPO~\cite{janner2019mbpo} (which itself uses SAC~\cite{haarnoja2018soft} internally), similar to MOPO. MBPO is a model-based RL algorithm that augments  Additional implementation details of MABE and hyperparameters are provided in the Appendix.

\paragraph{Experiments in D4RL offline RL benchmark tasks} Our first goal is to study the performance of MABE on the widely studied D4RL~\cite{fu2020d4rl} benchmark. We consider a total of nine domains involving three simulated locomotion tasks and three datasets per task: medium, medium-replay (or mixed), and medium-expert. The medium dataset is collected with partially trained SAC agent, the mixed dataset is the entire replay buffer of a SAC agent throughout training, and the medium-expert is a mix between trajectories from the medium dataset and an expert policy. These represent three distinct types of imperfect data - one imperfect policy, many changing policies, and a mixture of expert and suboptimal policies respectively. We compare our method to published leading offline RL algorithms which include: (a) MOReL~\cite{Kidambi-MOReL-20} and MOPO~\cite{yu20mopo} -- model-based algorithms that rely on uncertainty quantification; (b) CQL~\cite{kumar20cql}, a model-free algorithm that learns a conservative Q-function, and (c) BRAC-v~\cite{brac2019wu}, which regularizes a model-free actor-critic algorithm with an unweighted (or equally-weighted) behavioral prior. Please see appendix for more details.

Evaluation scores on D4RL are shown in Table~\ref{table:main}. We find that MABE achieves the highest score on the majority (7 out of 9) environments as well as the highest average score of $77.5$. Crucially, MABE's performance is robust across the three dataset types, achieving a leading score on at least 2 out 3 environments for each dataset. Finally, we note that MABE substantially outperforms its two most directly competing baselines: MOPO, an uncertainty-based MBRL method; and BRAC-v, a model-free method with explicit behavioral prior regularization. This suggests that a combination of MBRL and behavioral priors can substantially benefit offline RL. 

\label{section:experiments}

\begin{table}[h!]%
\caption{Normalized scores for the D4RL environments we consider. Scores for MABE are calculated from the average scores of the last 10 evaluation steps, over 3 seeds. Baseline results for MOPO, MOReL, and CQL are reproduced from their respective papers. SAC, BC, and BRAC-v numbers are reproduced from Fu et al.~\cite{fu2020d4rl}. We observe that MABE either matches or outperforms prior methods in a majority of the tasks, and achieves the highest average score.}
\begin{center}
\resizebox{\columnwidth}{!}{
\begin{tabular}{c|c|c|c|c|c|c|c|c}
\toprule
\textbf{Dataset} & \textbf{Environment} & \textbf{BC} & \textbf{MABE (ours)} & \textbf{MOPO} & \textbf{MOReL} & \textbf{SAC} & \textbf{CQL} &  \textbf{BRAC-v} \\
\midrule
medium & 
halfcheetah &
36.1 &
\textbf{46.8} $\pm$ 0.8 &
42.3 $\pm$ 1.6 &
42.1 &
-4.3  &
44.4 &
45.5
\\
medium &
hopper &
29.0 &
\textbf{94.1} $\pm$ 5.8 &
28.0 $\pm$ 12.4 &
\textbf{95.4} &
0.8 &
58.0 &
32.3
\\
medium &
walker2d &
6.6 &
65.7 $\pm$ 8.5 &
17.8 $\pm$ 19.3 &
77.8 & 
0.9 & 
79.2 & 
\textbf{81.3}
\\
med-replay &
halfcheetah &
38.4 &
\textbf{53.5} $\pm$ 0.5 &
\textbf{53.1} $\pm$ 2.0 &
40.2 &
-2.4 &
46.2 &
45.9
\\
med-replay &
hopper &
11.8 &
71.7 $\pm$ 12.5 &
67.5 $\pm$ 24.7 &
\textbf{93.6} &
1.9 &
48.6 &
0.9
\\
med-replay &
walker2d &
11.3 &
\textbf{51.0} $\pm$ 2.4 &
39.0 $\pm$ 9.6 &
49.8 &
3.5 &
26.7 &
0.8
\\
med-expert &
halfcheetah &
35.8 &
\textbf{100.6} $\pm$ 1.3 &
63.3 $\pm$ 38.0 &
53.3 &
1.8 &
62.4 &
45.3
\\
med-expert &
hopper &
111.9 &
\textbf{110.5} $\pm$ 0.8 &
23.7 $\pm$ 6.0 &
108.7 &
1.6 &
\textbf{111.0} &
0.8
\\
med-expert &
walker2d &
6.4 &
\textbf{103.3} $\pm$ 1.3 &
44.6 $\pm$ 12.9 &
95.6 &
-0.1 &
98.7 &
66.6 \\
\midrule 
Average & Average & 31.7 & \textbf{77.5} & 42.1 & 72.9 & 0.4 & 63.9 & 35.5
\\
\bottomrule
\end{tabular}
}
\label{table:main}

\vspace{-0.2in}
\end{center}
\end{table}

In the remainder of this section, we investigate in detail why MABE performs well and what new capabilities are enabled by MABE.

\begin{figure}[b!]
    \centering
    \includegraphics[height=4.5cm, width=\textwidth]{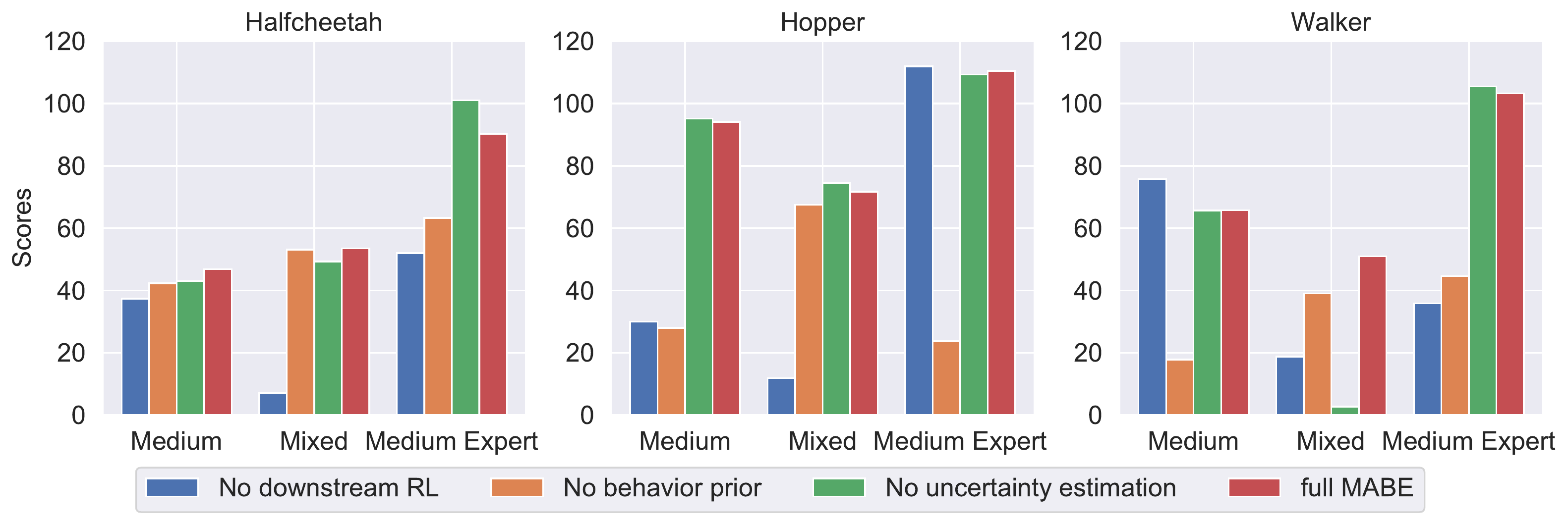}
    \vspace*{-10pt}
        \caption{Ablation over the three components of MABE. In most environments, the best performance is achieved from having all three components of MABE, but the component that gives the biggest boost in performance on average is the behavior prior. In most environments, uncertainty estimation does not materially affect the policy's performance. }
    \label{fig:ablation_components}
\end{figure}

\paragraph{Which components of MABE contribute most to performance?}
\label{ablations} MABE consists of several components that each play a part in the final agent. The full MABE algorithm consists of three components: (a) adaptive behavioral prior regularization; (b) policy learning (improvement) using model-based RL, and (b) the optional use of uncertainty quantification through model ensembles~\cite{chua18pets, RajeswaranGameMBRL, Kidambi-MOReL-20, yu20mopo} to incorporate additional conservatism. In this ablation study, we investigate the importance of each of these components by removing one while keeping all others fixed. 
Results shown in \autoref{fig:ablation_components}, indicate that RL and behavioral priors are the largest contributors to MABE's performance, while the optional uncertainty penalty only incrementally improves the final policies. Removing the uncertainty penalty leads to an observable drop in performance in only 2 out of the 9 environments. In contrast, removing behavioral priors drops performance in 8 environments, and removing RL drops performance in 7. Aggregated across the datasets, we find that removing behavioral priors results and RL result in a $41\%$ and $48\%$ drop in performance respectively. At the same time, removing uncertainty estimation only marginally degrades MABE performance by $9\%$. This suggests that MABE has the potential to find wider applicability, especially in situations where uncertainty estimation can be difficult, but can also benefit from uncertainty estimation where available.

In Figure~\ref{fig:ablation_components}, no downstream RL refers to the direct use of the adaptive behavioral prior, without any finetuning with MBRL. This can be viewed as a baseline inspired by imitation learning. The ablation study of no-behavioral prior corresponds to MOPO and incorporates conservatism through the use of uncertainty estimation. The no uncertainty estimation ablation utilizes adaptive behavior prior regularization to incorporate conservatism when learning the policy using MBRL. This utilizes all the components of the full MABE algorithm except the optional uncertainty-based reward penalties. Finally, the full MABE algorithm uses all the three aformentioned components of behavioral priors, policy learning with MBRL, and additional conservatism through uncertainty penalized rewards.

\begin{figure}[t!]
  \centering
  \includegraphics[width=\textwidth]{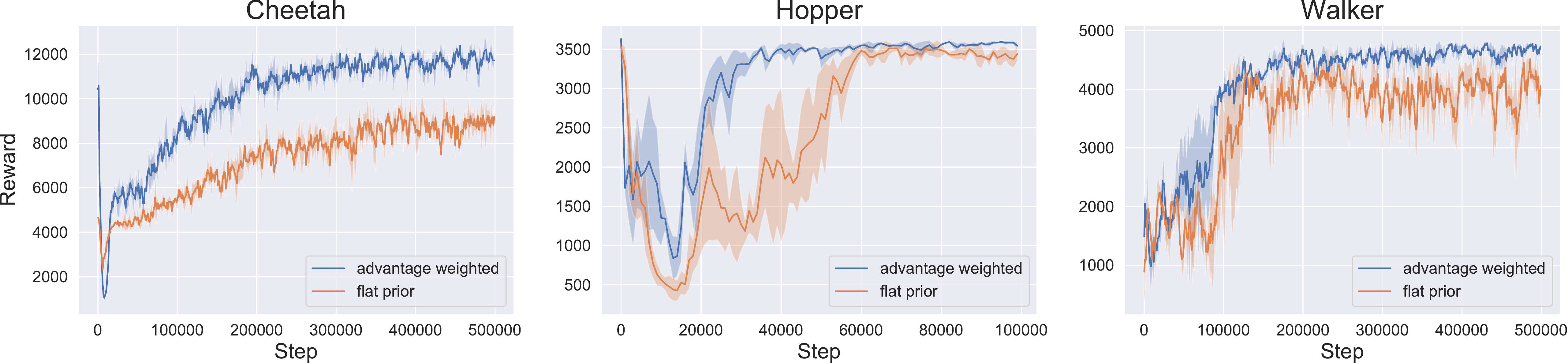}
  \vspace*{-10pt}
  \caption{Comparison of MABE run with an advantage weighted behavioral as well as a non-weighted ``flat"  prior. We find that advantage-weighing improves the stability and performance of the policy.}
  \vspace*{-15pt}
  \label{fig:prior_fig}
\end{figure}

\paragraph{Weighted vs Unweighted Behavioral Prior Regularization} Finally, we ablate the importance of adaptive or weighted behavioral priors as used in MABE. In particular, we compare MABE with the unweighted behavioral prior in Eq.~\ref{eq:unweighted_prior} against the full MABE algorithm that uses the adaptive prior in Eq.~\ref{eq:adaptive_prior}. We show learning curves for MABE trained with the two priors in Figure~\ref{fig:prior_fig} and find that adaptive priors help with training stability as well as asymptotic performance.

\begin{figure}[b!]
  \centering
  \includegraphics[width=0.85\textwidth]{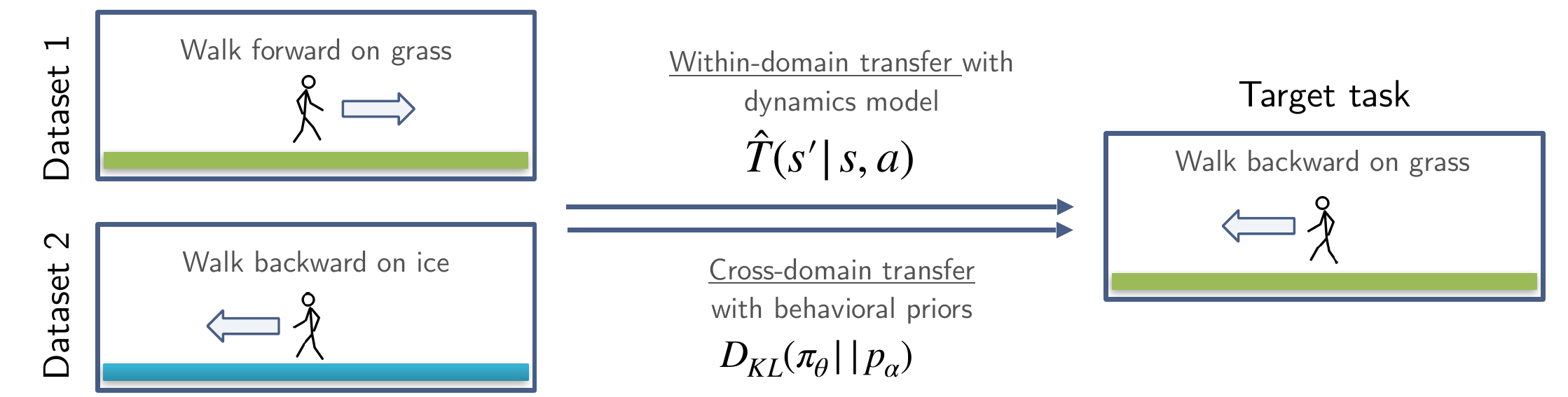}

  \caption{A schematic visualization of MABE's domain transfer capabilities. In prior work it was shown that offline MBRL is capable of in-domain generalization to new tasks~\cite{yu20mopo}. Here, we investigate cross-domain transfer capabilities of MABE and MOPO. Given multiple datasets with different dynamics and behaviors, can we generalize to a new task in the target domain that was not present in the offline data for the target domain? We hypothesize that behaviors are transferable across domains even if the dynamics are different. }
  \vspace*{-10pt}
  \label{fig:transfer}
\end{figure}

\paragraph{Cross-domain and cross-task generalization capability of MABE} A unique capability enabled by the use of behavioral priors is the possibility of transferring behaviors from one environment (or domain) to another. Prior work has explored the use of offline datasets and RL to acquire new behaviors in the same environment. For example, Yu et al.~\cite{yu20mopo} demonstrates that offline RL using a dataset that primarily consists of an agent walking forward can be used to learn a jumping behavior. In contrast, we seek for the agent to learn the same behavior but in a different environmental condition. This is particularly useful in robotics applications, like for instance home robots that operate in kitchens. While the environmental scene and physical dynamics would vary across different kitchens depending on the types of cabinets, stoves, plates, floor etc. we would often want to robot to exhibit similar behaviors in different kitchens like loading plates in a dishwasher. By utilizing behavioral priors that can potentially capture the core concepts of manipulation like force closure for grasping, robots can learn to become competent quickly in the home of a target user.

\begin{wrapfigure}{r}{0.5\textwidth}
     \centering
     \vspace{-2mm}
       \includegraphics[width=.45\textwidth]{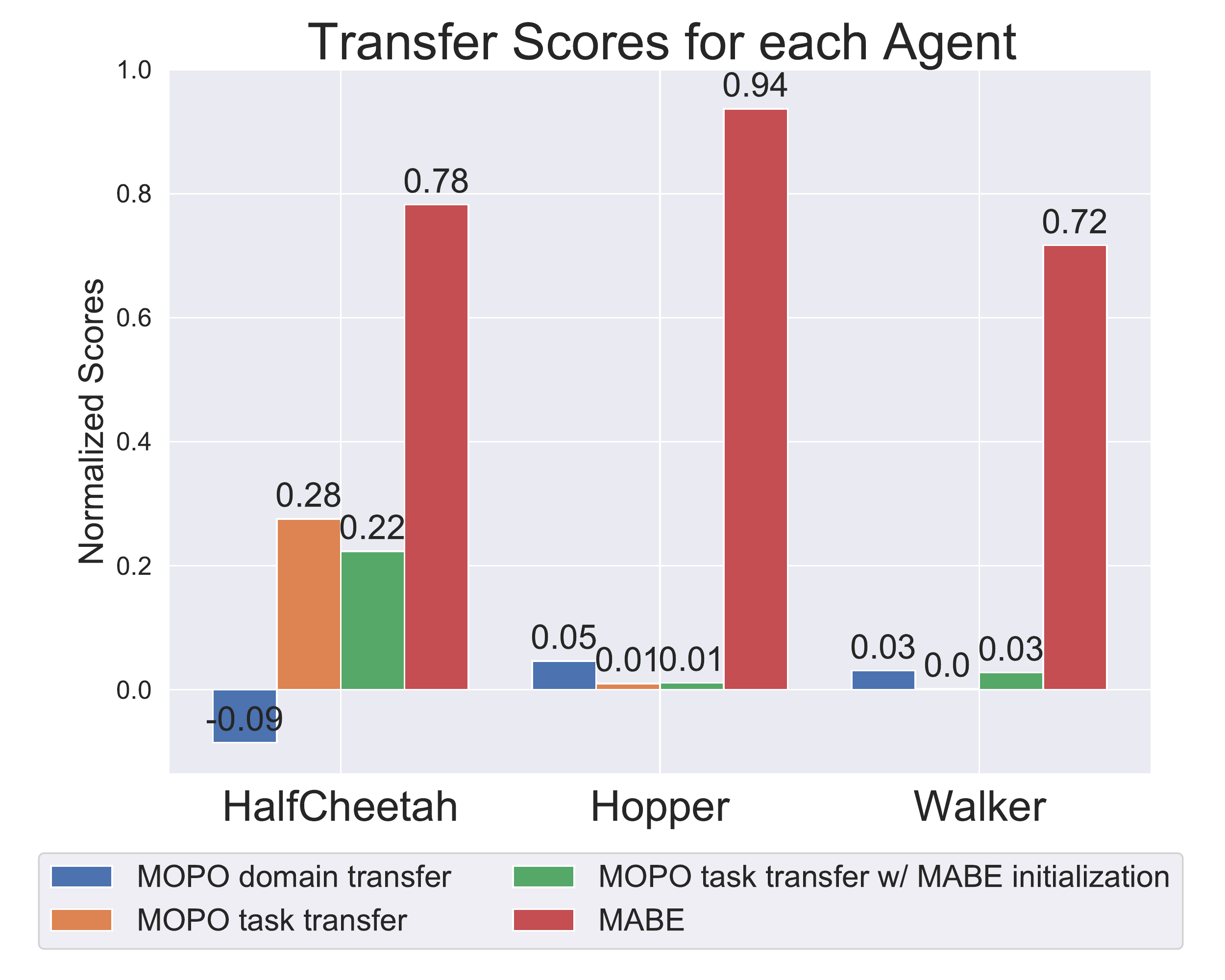}
     \caption{Experiments comparing MABE domain transfer to MOPO domain transfer and MOPO task transfer. MABE is the only offline MBRL approach that is able to successfully transfer behaviors across domains. }
    \vspace{-5mm}
     \label{fig:transfer_results}
 \end{wrapfigure}

To test the generalization capabilities of MABE, we setup the following simple experiment. We use simulated locomotion agents (Hopper, Walker, HalfCheetah), and collect two datasets: $\mathcal D_1$ containing medium-replay forward walking data in normal terrain; and $\mathcal D_2$ containing expert backwards walking data in low friction terrain intended to simulate ice. In this behavior transfer test, we use these two datasets to train an agent to run backward on normal terrain. A schematic illustration of our setting can be found in Figure~\ref{fig:transfer}.
In our experiments, we consider the following approaches: (i)  {\it task transfer only} where we use the forwards walking dataset to learn a backwards walking policy using offline MBRL. (ii)~{\it domain transfer only} where we train a policy in source domain and directly deploy it in the target domain. (iii)~{\it task transfer with behavior initialization} where we initialize the task transfer approach with the adaptive behavioral prior; (iv)~{\it task + domain transfer with MABE} where we run MABE using the dataset corresponding to the target dynamics $(\mathcal{D}_1)$ and behavioral prior corresponding to the desired behavior $(\mathcal{D}_2)$. We show the resulting expert normalized scores in Figure~\ref{fig:transfer_results} and find that MABE is the only algorithm that is able to successfully solve the target task through cross-domain behavior transfer. This suggests that dynamics models and behavioral priors are complementary and can be used to acquire a wide range of behaviors from offline data using domain and task transfer.

\section{Related Work}

Our method, MABE, is at the intersection of model-based reinforcement learning, offline reinforcement learning, and behavioral prior regularization. There are a number of related algorithms that utilize dynamics models or behavioral priors in the context of offline RL \cite{yu20mopo,brac2019wu,MatsushimaBREMEN,awr2019peng,Siegel2020KeepDW}, which we describe in Table~\ref{table:algos} with a comprehensive overview. While MABE is similar to prior work, our primary contribution is identifying a unique mixture of components that enable robust offline RL on the D4RL benchmark. Recently, concurrent work COMBO \cite{yu2021combo} has also investigated an uncertainty-free approach to offline MBRL. The difference is that COMBO combines offline MBRL with conservative Q-functions whereas MABE utilizes adaptive behavioral priors, which helps with cross domain generalization capability as demonstrated in Section~\ref{section:experiments}.

\noindent{\bf Model-based Reinforcement Learning}: Reinforcement learning algorithms can be broadly classified into model-based and model-free categories. Model-based reinforcement learning (MBRL) algorithms build an explicit dynamics model of the environment for use with policy search. Model-based approaches can be further categorized into Dyna-style algorithms, policy search with temporal backpropagation, and shooting methods. In dyna-style approaches \cite{sutton1990integrated,sutton1991dyna,sutton1991planning}, interactions with the environment are used to update the dynamics model and the RL policy is trained on synthetic rollouts from the dynamics model, often using a model-free RL algorithm like policy gradients or actor-critic. Some representative examples of Dyna-style algorithms include MBPO~\cite{janner2019mbpo}, ME-TRPO~\cite{kurutach2018model}, PAL/MAL~\cite{RajeswaranGameMBRL}, and Dreamer~\cite{hafner2019dream}. Policy search with temporal backpropagation and differential dynamic programming methods~\cite{Rosenbrock1972DDP, Deisenroth2011pilco, Heess2015SVG, tassa12ilqg, Todorov2005iLQG} utilize gradients through the model to help compute the policy gradient. Shooting methods~\cite{chua18pets, hafner2018learning, Williams2017MPPI, Nagabandi2019PDDM, POLO} extract an implicit policy from the learned model by performing real-time planning using the learned model. For simplicity and to build on prior work in the area of offline RL, we implemented MABE with MBPO, a Dyna-style algorithm. However, MABE can in principle be implemented with any MBRL algorithm.

\noindent{\bf Offline Reinforcement Learning}: Offline RL~\cite{levine2020offlinerlsurvey} has recently received much attention due to its potential for applicability in a wide range of applications, and consequently many algorithms have been developed recently. Among them include importance sampling based algorithms~\cite{Liu2020ProvablyGB, LiuSAB19, SwaminathanJ15}, dynamic programming and actor-critic based algorithms~\cite{brac2019wu, fujimoto2018addressing, Agarwal2020AnOP, Siegel2020KeepDW, kumar20cql}, and model-based algorithms~\cite{Kidambi-MOReL-20, yu20mopo, MatsushimaBREMEN, ArgensonMBOP}. These algorithms are primarily evaluated using recently proposed benchmarks including D4RL~\cite{fu2020d4rl}, Atari~\cite{Agarwal2020AnOP, bellemare2013arcade} and RL-Unplugged~\cite{Gulcehre2020RLUA}. We outline the contrasts between MABE and prior work in the remainder of the section.

\noindent {\bf Relationship to prior offline MBRL algorithms} In terms of the policy learning, our work is closest to prior offline MBRL algorithms -- MOPO~\cite{yu20mopo} and MOReL~\cite{Kidambi-MOReL-20}, which rely on uncertainty quantification to estimate model prediction error to incorporate conservatism. In contrast, MABE can benefit from uncertainty estimation, but even in its absence demonstrates strong performance and thus has wider applicability. BREMEN~\cite{MatsushimaBREMEN} is another MBRL algorithm that was primarily developed for a different setting of deployment efficient RL but can be re-purposed for offline RL. Like MABE, it uses a behavioral prior instead of uncertainty driven conservatism. However, it uses an unweighted behavioral prior and performs only a small number of policy updates with implicit KL regularization. As a result, it may not benefit from the full potential of policy learning for many iterations with an explicit KL regularization. Furthermore, in our experiments (Section~\ref{section:experiments}), we find that adaptive behavioral prior helps learning stability and improves asymptotic performance.

\begin{table}[t]
\caption{A comparison between MABE and prior algorithms.  }
\vspace*{-15pt}
\label{table:algos}
\begin{center}
\resizebox{\columnwidth}{!}{
\begin{tabular}{l|llllll}
\toprule
& MABE 
(ours) & MOPO/MOReL & BRAC-v~\cite{brac2019wu} & BREMEN~\cite{MatsushimaBREMEN} & ABM~\cite{Siegel2020KeepDW} & AWR~\cite{awr2019peng} \\ \midrule
Model-Based  & Yes & Yes & No & Yes & No & No\\ 
Behavior Prior & Adaptive & None & Unweighted & Unweighted & Adaptive & Adaptive \\ 
Policy Regularization & Explicit KL & None & Explicit KL & Implicit KL & Implicit KL & Implicit KL \\  
Policy Optimization & SAC & SAC/NPG & SAC & TRPO & MPO~\cite{Abdolmaleki2018MPO} & Imitation  \\  
Uncertainty & Optional & Yes & No &  No & No & No \\  \bottomrule
\end{tabular}}
\end{center}
\vspace*{-15pt}
\end{table}

\noindent{\bf Relationship to prior work with behavioral priors}: An alternate class of offline RL algorithms incorporate conservatism to prevent over-fitting by regularizing the policy learning towards a behavioral prior. Some representative algorithms are BRAC~\cite{brac2019wu}, ABM~\cite{Siegel2020KeepDW}, and AWR~\cite{awr2019peng}, which are all model-free algorithms. Among these, BRAC uses an unweighted behavioral prior and learns the policy using an actor-critic algorithm like SAC~\cite{haarnoja2018soft}. AWR was primarily developed for online RL but can be re-purposed for offline RL. It is analogous to our learning of adaptive behavioral prior, but without any RL based fine-tuning. In our ablation experiments, we find that RL finetuning significantly improves the performance of MABE. ABM learns an adaptive behavior prior similar to MABE, but learns the policy using the model-free MPO algorithm. In contrast, model-based algorithms that augment training data with model-generated rollouts can unlock better generalization capabilities, including to new tasks. In an alternate line of work, behavioral priors have also been used for skill extraction to enable long-horizon tasks~\cite{pertsch2020spirl} or structured exploration strategies~\cite{singh2021parrot}. Finally, we also note that the concurrently developed Decision Transformer~\cite{chen2021decisiontransformer} learns a return-conditioned behavior model and generates actions by conditioning on high desired reward. In contrast, MABE does not condition on returns and instead uses an RL algorithm for policy improvement.

In summary, we note that MABE presents a novel combination of MBRL and adaptive behavioral priors for offline RL. Through this combination, MABE can serve as an attractive choice for uncertainty-free offline MRBL. MABE also achieves leading performance relative to prior model-based and model-free approaches on the D4RL benchmark and demonstrates a strong ability to transfer behaviors across datasets from different domains.

\section{Broader Impacts and Limitations}
\label{sec:limitations}
Robust offline RL has the potential to make RL as widely applicable for  decision making problems as supervised learning is today for vision and language. Applications include domains where offline data is ample but exploration can be harmful such as controlling autonomous vehicles, digital assistants, and recommender systems. Negative potential impacts of MABE and RL algorithms more generally is the lack of explainability. Since MABE is simply optimizing a reward function while regularizing against a behavioral prior it can learn policies with undesired consequences that exploit the reward function. Future work on explainability of RL policies as well as constrained policy optimization could help alleviate these concerns. 
While we extensively evaluate our method using D4RL benchmark tasks, and also study cross-domain transfer, our experimental evaluation is in continuous control tasks. Although continuous control is representative of many applications in robotics, offline RL is a broad and vibrant field with applications involving language~\cite{jaques2019way, ZhouSRE17} and visual modalities~\cite{Agarwal2020AnOP, Rafailov2020LOMPO, hafner2019dream}. We hope to extend MABE to different offline RL tasks and high-dimensional observation modalities in future work. 

\section*{Acknowledgments}

This work was supported by Berkeley Deep Drive. Part of this work was completed when Aravind Rajeswaran was at the University of Washington, where he was supported through the J.P. Morgan PhD Fellowship in AI (2020-21). The authors thank Kevin Lu and Justin Fu for help with setting up the D4RL benchmark tasks. All content represents the opinion of the authors, which is not necessarily shared or endorsed by their respective employers and/or sponsors.

\bibliography{main}{}
\bibliographystyle{unsrt}
\newpage
\appendix

\section{Environments}
In our experiments, we use offline datasets from D4RL \cite{fu2020d4rl} for environments from OpenAI gym's \cite{brockman2016openai} MuJoCo continuous control tasks \cite{6386109}. We look at three locomotion agents shown in \autoref{fig:mujoco}: HalfCheetah, Hopper, and Walker2d, which are all tasked with moving forward as fast as possible. For each agent, we look at three types of datasets: 
\begin{enumerate}
\item \textbf{Medium}: Approximately 1 million transitions collected from a partially trained SAC agent
\item \textbf{Mixed}: Approximately 100000 transitions collected from the entire replay buffer of a SAC agent throughout training
\item \textbf{Medium-expert}: Approximately 2 million transitions consisting of half medium samples (collected from a partially trained SAC agent) and half expert samples, which are collected from a fully trained SAC agent. 

\end{enumerate}
We don't evaluate on random datasets, which are collected with a random policy for two reasons. First, the actions in these datasets are completely random and behavioral priors are not expected to be helpful since the behaviors are random. Instead we are more interested in evaluating performance on offline datasets with some, even if minimal, structure. Second, we argue that completely random data is a somewhat contrived benchmark. Datasets used to solve real-world problems in robotics, such as autonomous vehicle navigation, locomotion, and manipulation are likely to have some sort of structure.

\section{Baselines}
We compare against several leading model-based and model-free offline RL baselines on the D4RL dataset. 
\begin{enumerate}

\item {\bf MOPO}: MOPO \cite{yu20mopo} is an uncertainty-based offline MBRL algorithm. MOPO uses MBPO \cite{janner2019mbpo}, an off-policy Dyna-style RL  algorithm where a replay buffer is populated with synthetic samples from a learned dynamics model and used to train an Soft Actor Critic (SAC) \cite{haarnoja2018soft} agent. MOPO build on MBPO by penalizing the reward experienced by an agent with a penalty proportional to the prediction uncertainty of the dynamics model. MABE is also built on top of MBPO and thus MOPO is the most directly competing baseline.
    \item {\bf MOReL}: MOReL \cite{Kidambi-MOReL-20} is also an uncertainty-based offline MBRL algorithm. The primary difference between MOReL and MOPO is that MOReL uses an on-policy algorithm, TRPO \cite{schulman2015trpo}, as its backbone. Otherwise, MOPO and MOReL are similar - both penalize the reward with a term proportional to the forward model uncertainty. The performance differences between MOPO and MOReL on D4RL are mainly due to the performance of the backbone algorithm,  SAC and TRPO respectively. SAC outperforms TRPO on the mujoco Cheetah environment while TRPO outperforms TRPO in the Hopper environment, and these differences are also evident in the offline RL results for MOPO and MOReL.
    \item {\bf CQL}: Conservative Q-Learning (CQL) \cite{kumar20cql} is a leading offline model-free baselines.  CQL learns Q-functions so that the expected value of a policy under the learned Q-function is a lower-bound of the true policy value. CQL modifies the standard Bellman error with a term that minimizes the Q-function under the policy distribution while maximizing it under the offline data distribution. CQL does not leverage behavioral priors.
    \item {\bf BRAC-v}: BRAC-v is another leading model-free  RL algorithm that utilizes behavioral priors to learn a conservative policy. BRAC-v is the model-free algorithm most similar to MABE. Like MABE, BRAC-v learns a behavioral prior by fitting a Gaussian distribution to the offline data and regularizing a Gaussian evaluation policy with respect to the behavioral data. Unlike MABE, BRAC-v does not weigh the behavioral prior with the advantage and instead treats all data points equally regardless of the reward achieved.

\end{enumerate}

Additionally, we include comparisons to naive behavior cloning and offline SAC.

\section{Experiment Details}

\subsection{Advantage-Weighted Behavioral Prior}
First, to learn the advantages for each datapoint in dataset, we fit a Q-function to the offline dataset. We train until the loss no longer increases any further, then use this Q-function to assign Q-values to each datapoint. We normalize these Q-values by dividing each value by the maximum Q-value assigned to any datapoint.

We train our behavioral prior using a negative log likelihood loss. We weight the loss from each datapoint by the exponentiated normalized Q-values obtained from our learned Q-function. During training, we do a 90-10 train-validation split and stop training when the validation loss stops decreasing.

One note is that for halfcheetah medium-expert, we found that a more simple weighing scheme led to better results. Rather than fitting a Q-function, we weighed datapoints by the final total reward of their trajectory instead. For all other environments, we found that weighing by the Q-function worked better or approximately the same.

\subsection{Hyperparameters}

Because we built off of MOPO \cite{yu20mopo}, we use the same MOPO-specific hyperparameters for the MOPO hyperparameters of the rollout length $h$ and penalty coefficient $\lambda$. We refer you to the MOPO Appendix for these values. We additionally use the MOPO architecture and training method for our dynamics model ensemble. For the dynamics model, we train an ensemble of 7 dynamics models and choose the 5 best models based on their prediction error to use while training our offline SAC agent. 

For our policy network, we learn a Gaussian two-head network with 2 hidden layers with 256 hidden units, and two separate linear output layers outputting the mean and log standard deviation of the next action. For our Q networks, we use an architecture of 3 feed-forward layers of 256 hidden units each. Our behavioral prior has the same architecture as our policy network.

Our main hyperparameter for our method is the target KL divergence $\delta$. For our hyperparameter search, we defaulted on a low target divergence for the medium-expert datasets ($\delta = 0.1$), and we performed a grid search for the medium and medium-replay environments, because we found that the different agents required different target divergences based on their dataset composition. The full list of target divergences used can be found in Table~\ref{table:divergence}

\begin{table}[t]
\vspace*{-15pt}
\label{table:divergence}
\begin{center}
\resizebox{8cm}{!}{
\begin{tabular}{l|l|l}
\toprule
Dataset Type & Environment & Target Divergence $\delta$ \\ \midrule
medium & halfcheetah & 100 \\
medium & hopper & 0.75 \\
medium & walker2d & 1 \\
mixed & halfcheetah & 40 \\
mixed & hopper & 5 \\
mixed & walker2d & 20 \\
medium-replay & halfcheetah & 0.1 \\
medium-replay & hopper & 0.1 \\
medium-replay & walker2d & 0.1
\\
  \bottomrule
\end{tabular}}
\end{center}
\caption{Table of target divergences used in MABE per environment}

\vspace*{-15pt}
\end{table}

\section{Compute Resources and Assets Used} 

\paragraph{Compute Resources} Experiments for our main suite of results were run on  GPUs using a machine with eight Quadro RTX 6000. However, only one GPU is required for four concurrent experiments, so our main experiments used approximately 1080 GPU hours (including all seeds).

\paragraph{Assets Used} In this work we used the D4RL Offline RL Benchmark for evaluation \cite{fu2020d4rl} which has an Apache License 2.0. We build our code off of logic from MOPO \cite{yu20mopo},  which is distributed under a MIT License. We built our final codebase off of a \href{https://github.com/jxu43/replication-mbpo}{PyTorch replication codebase of MBPO}. From this codebase, we ported over MOPO logic. Additionally, we train our dynamics models in the MOPO official codebase for fair comparison against MOPO. For our baselines, we report results for MOPO, BC, SAC, and BRAC-v from \cite{yu20mopo}, MOReL from \cite{Kidambi-MOReL-20}, and CQL from \cite{kumar20cql}.

\end{document}